\documentclass[10pt, a4paper]{article}

\usepackage[final]{lrec2026} 
\usepackage[utf8]{inputenc}
\usepackage[T1]{fontenc}

\title{OnCoCo 1.0: A Public Dataset for Fine-Grained Message Classification in Online Counseling Conversations}

\name{\parbox{\textwidth}{\centering 
      Jens Albrecht, Robert Lehmann, Aleksandra Poltermann, \\
      Eric Rudolph, Philipp Steigerwald, and Mara Stieler\vspace{0.2cm}}}
\address{Technische Hochschule Georg Simon Ohm \\
         Kesslerplatz 12, 90489 Nuremberg, Germany}

\abstract{
This paper presents OnCoCo 1.0, a new public dataset for fine-grained message classification in online counseling.
It is based on a new, integrative system of categories, designed to improve the automated analysis of psychosocial online counseling conversations. Existing category systems, predominantly based on Motivational Interviewing (MI), are limited by their narrow focus and dependence on datasets derived mainly from face-to-face counseling. This limits the detailed examination of textual counseling conversations. In response, we developed a comprehensive new coding scheme that differentiates between 38 types of counselor and 28 types of client utterances, and created a labeled dataset consisting of about 2.800 messages from counseling conversations. We fine-tuned several models on our dataset to demonstrate its applicability. The data and models are publicly available to researchers and practitioners. Thus, our work contributes a new type of fine-grained conversational resource to the language resources community, extending existing datasets for social and mental-health dialogue analysis.
\\ \newline \Keywords{Classification, Counseling, Conversations, Dataset, Fine-grained annotation}
}

\begin{document}
\maketitleabstract
\titlespacing*{\paragraph}{0pt}{1ex plus 0.2ex minus .1ex}{1em}

\section{Introduction}

Psychosocial counseling covers a wide range of human problems. The WHO 2021 report states that 1 in 8 people worldwide is struggling with mental health problems and disorders \cite{mentalHealthAtlas2021}. In other areas, such as pregnancy, parenting, or debt counseling, the demand for advisory services is similarly increasing \cite{nhsRecordNumbersWomen2024}.

Nowadays, people can seek help on a wide range of psychological and social topics through web forums and dedicated text-based counseling platforms. \textit{Online counseling} is provided on these platforms by, for example, medical professionals, psychologists, or social workers who have undergone specialized training in this method \cite{krausOnlineCounselingHandbook2010}. It is characterized by an inherently asymmetric social relationship: counselors and clients occupy different social roles in the sense of interaction role‑taking, where participants continuously negotiate positions such as "helper" and "help‑seeker". A trained counselor guides the interaction, while the client seeks support under conditions of vulnerability and limited contextual cues. By providing fine-grained role-specific labels for both sides, our dataset OnCoCo 1.0 enables the study of social dynamics in supportive interactions, e.g. how guidance, empathy, and resource activation unfold across counselor-client turns.

Currently, there is an intensifying debate regarding the potential of AI systems to assume the role of online counselors or even therapists \cite{hatchWhenELIZAMeets2025}. Psychotherapy research \cite{laskaExpandingLensEvidencebased2014} and studies on psychosocial counseling \cite{grawePsychologischeTherapieGottingen2000} emphasize that key mechanisms of change are strongly shaped by the content conveyed in counseling interactions. To gain a more nuanced understanding of whether and how online counseling exerts its effects, it is therefore essential to analyze the specific thematic elements (e.g. agreement on consulting goals or creation of motivation) present within a counseling exchange \cite{grandeitUsingBERTQualitative2020}. Particularly, in German-language research on online counseling, these are characterized by a systematic approach using inductive or deductive coding \cite{mayringQualitativeContentAnalysis2015}. 

So far, data sets specifically for online counseling have been hardly publicly available, mainly due to privacy-related restrictions or reasons of data protection. In addition, existing category systems for automated content analysis focus mainly on specific counseling approaches, such as Motivational Interviewing (MI) \cite{millerManualMotivationalInterviewing2008, moyersMotivationalInterviewingTreatment2016, xiaoBehavioralCodingTherapist2016, wuAnnoMIDatasetExpertAnnotated2022}. These systems often focus on face-to-face settings \cite{malhotraSpeakerTimeawareJoint2022, wuAnnoMIDatasetExpertAnnotated2022}, which does not capture the full details of relevant aspects in online counseling where text is the only modality. 
A central challenge is that online counseling is quite different to face-to-face counseling in terms of methodology. To compensate for the lack of auditory and visual cues, it employs specific reading and writing techniques that deviate significantly from those used in face-to-face conversations.

\paragraph{Contribution:}
With this publication we introduce OnCoCo 1.0 (\textit{On}line \textit{Co}unseling \textit{Co}nversations), a new bi-lingual dataset (German and English) for rich content analysis in psychosocial online counseling. Our dataset extends current conversational corpora by providing a detailed and ethically curated dataset for bilingual counseling contexts. 
For the classification, we developed a novel and fine-grained category system, using methods of qualitative social research \cite{mayringQualitativeContentAnalysis2015} and grounded in theoretical models of online counseling (e.g., \cite{grawePsychologischeTherapieGottingen2000}). It is composed of 38 categories for counselor messages and 28 categories for client messages. Furthermore, 2.778 messages were manually created and curated by professional experts and specifically advised students. We fine-tuned several models based on this data to demonstrate it's applicability. The best model reaches an accuracy of 80\% and an F1 macro score of about 0.78 which is comparable to human performance on this kind of data.
The developed category system extends previous work in the sense that it is not limited to just one method within counseling, e.g. MI, but corresponds to the special claim of online counseling - i.e. the mixture of different methods.
Hence, our system and the corresponding models are suitable for analyzing all types of texts during counseling or therapy sessions. 

\paragraph{Availability and License:}
The OnCoCo 1.0 dataset, including documentation, metadata and limitations, as well as ready-to-use fine-tuned models are publicly available for research and educational purposes under a CC BY 4.0 license. The data and models can be accessed at \url{https://huggingface.co/datasets/th-nuernberg/OnCoCoV1}.
A full metadata description is also provided via the LRE Map.

\section{Relevance of our Research}

The ability to analyze unlimited amounts of conversations with a rich system of categories offers completely new possibilities for online counseling research. Fine-grained patterns can be collected and transferred in complex and comprehensive statistical models. 
This significantly expands our knowledge to relevant aspects of successful counseling. 

\paragraph{Social Interaction Pattern Analysis:}
The fine-grained and role-specific coding scheme enables detailed analyses of counseling interaction patterns and social dynamics.
For example, it supports (i) \textit{strategy--response} analyses by studying transitions between counselor categories (e.g., problem clarification vs.\ motivation building) and subsequent client categories (e.g., engagement, resistance, or requests for help), and (ii) \textit{conversation stage} analyses by comparing category distributions between early and late turns.
These analyses are difficult to conduct with coarse tag sets and 
illustrate how the resource can be used in qualitative and mixed-methods research.

\paragraph{Quality Improvement:}
Classification models trained on our dataset can be used to identify patterns and success factors for online counseling, such as problem solving, relationship-building, interventions etc. (see Table~\ref{tab:CategoriesSamples}). 
Text analysis is particularly well suited to explore the micro-level of counseling relationships such as interaction patterns, emotional dynamics, and conversational flow. 
In face-to-face counseling, conversation and text analyses of recorded sessions have already provided valuable insights into interaction processes and relationship-building
\cite{hanckenBeziehungsgestaltungSozialenArbeit2023}. 
These approaches can be adapted to the context of online counseling. 
AI-supported analysis also enables the examination of large volumes of counseling transcripts, helping to detect and quantify latent patterns and meaningful structures. This can assist professionals in reflecting on their methods and incorporating successful interaction strategies into their practice or training.  
Beyond individual counselors, supervisors can also use these insights for case reviews, team reflection and structured monitoring processes to improve quality assurance.

\paragraph{Education:}
Counselor education can be supported by providing valuable insights and resources.
For instance, trainees can receive real-time feedback on their performance by having their counseling sessions automatically analyzed by the model. 
It could highlight areas in which the counselor needs to work on developing empathy strategies, setting goals, intervention techniques, or activating resources. 
This feedback can be included into the educational process to direct future development possibilities. Additionally, trainees can learn to identify patterns in interactions between counselors and clients. 
Since the model may categorize various aspects of a conversation, such as the counselor's use of motivational strategies, educators can pinpoint certain strengths and shortcomings in student performance. 
In addition to real-world sessions, the model can help to create simulated environments for counselor training. By generating potential interactions between the counselor and the client (or analyzing past simulated dialogues), trainees can practice identifying appropriate responses, intervention strategies, and emotional cues.

\paragraph{Detection of Common Issues:}
Emerging trends and prevalent issues, such as those that teenagers encountered during the COVID-19 pandemic, can be found with the aid of automated analysis. Early trend detection allows for prompt and focused intervention in social and mental health issues, which shortens the duration of these problems for impacted individuals and lessens their wider societal impact. 
\paragraph{Evaluation of Chatbots:} Studies based on our dataset enable a fine-grained evaluation of counseling interactions conducted by counseling chatbots. On this basis, AI-based counseling systems can first be assessed in terms of their counseling quality and subsequently optimized to reach the standard set by human counselors.

\paragraph{Efficient Resource Allocation:} In depth data analysis within counseling organizations can support better resource allocation in social work. For example, counselors can be directed to cases requiring urgent attention (e.g., high-risk situations such as suicide or child welfare concerns), thus optimizing the use of human resources and offering better support for clients in urgent need of help.




\section{Related Work}
\label{sec:RelatedWork}
In the context of online counseling, content analysis has been investigated in some work during the last decade.
\cite{althoffLargescaleAnalysisCounseling2016} developed models to evaluate conversation strategies such as adaptability, progress, and shifts of perspective based on an SMS-data that are not publicly available.

Several other publications are based on Motivational Interviewing (MI) \cite{millerMotivationalInterviewingPreparing2002}. 
The \textit{Motivational Interviewing Skill Code} (MISC) is a coding scheme designed to analyze both client and counselor behavior in MI sessions \cite{millerManualMotivationalInterviewing2008}. 
The MISC category system comprises a total of 28 categories (19 for counselor, 9 for clients).
\cite{xiaoBehavioralCodingTherapist2016} and \cite{caoObservingDialogueTherapy2019} applied machine learning to MISC, but used a simplified set of just 8 counselor and 2 client categories because the support for some categories in their dataset was considered too low. 

\cite{perez-rosasBuildingMotivationalInterviewing2016} describe the creation of a dataset based on the even simpler
\textit{MI Treatment Integrity} (MITI) coding system, which focuses solely on the counselor. 
Utterances are categorized into MI-adherent (e.g. simple/complex reflection, affirmation) and MI-non-adherent (e.g. direct persuasion or confrontation).
Their dataset contains a total of 22,719 coded utterances. Later on, it was used to analyze counseling conversations with regard to empathy and overall quality of counseling \cite{perez-rosasUnderstandingPredictingEmpathic2017, perez-rosasWhatMakesGood2019}. 
Although stated otherwise in the publications, this dataset is currently not available to the public.

\textit{Anno-MI}, in contrast, is a public dataset that includes transcribed therapy dialogues on MI annotated by professional therapists \cite{wuAnnoMIDatasetExpertAnnotated2022, wuCreationAnalysisEvaluation2023}. It consists of 133 videotaped and transcribed conversations with a total of over 9.699 utterances. The annotation scheme is, however, rather coarse, consisting of just the three major therapist behaviours: reflection (simple/complex), question (open/closed), and input (information, advice, giving options, and negotiation). The subcategories (in parentheses) were defined but not used for annotation because of low inter-coder agreement in a first test. Client messages were also differentiated into three categories, namely change, neutral, and sustain similar to MISC.

Based on this previous work on MI, \cite{cohenMotivationalInterviewingTranscripts2024} recently created a new public MI dataset consisting of 242 sessions with 15.627 total messages. 
Therapist behavior is classified into 11 categories based on MITI. 

The dataset \textit{ESConv} (Emotional Support Conversation) as well as a corresponding framework for the modeling of dialogues and corresponding categories were published by \cite{liuEmotionalSupportDialog2021}. The coding scheme puts more emphasis on the client side and distinguishes between five types of client problems, seven emotions, and a feedback score between one and five. Eight support strategies are annotated on the counselor side.
The publicly available dataset includes 1,053 conversations consisting of over 31,000 utterances.

The public \textit{HOPE} dataset \cite{malhotraSpeakerTimeawareJoint2022} contains approximately 12,900 statements from 212 therapy sessions in the field of cognitive behavioral therapy (CBT). For the annotation, a category system consisting of 12 categories, organized into three main categories, was developed. 


In addition, there exist general dialogue-act frameworks such as ISO~24617-2 \cite{bunt-etal-2020-iso} which can serve as a shared reference point for relating the task- and counseling-specific annotation schemes discussed above.

\paragraph{Conclusion}

Many existing datasets are not publicly accessible, and those that are often use overly coarse annotation schemes that prevent detailed analysis, or are based on a face-to-face counseling setting.
Furthermore, it is crucial to differentiate between \textit{counseling} (social) and \textit{therapy} (medical), as they use different methodological approaches. Despite these differences, there are numerous similarities that have influenced the subsequent heterogeneous selection of categories in our study.

\section{The Category System}\label{sec:CategorySystem}

To enable a structured and fine-grained analysis of online counseling conversations, we developed a hierarchical category system that captures diverse aspects of both client and counselor messages.

\renewcommand{\arraystretch}{1.3}

\begin{table*}[ht]
\centering
\sffamily\small
\scalebox{0.9}{%
\begin{tabular}{|l|l|l|c|c|}
\hline
\textbf{Role} & \textbf{Categories on Level 1} & \textbf{Categories on Level 2} & \textbf{\#Subcategories} & \textbf{Samples} \\ \hline
Counselor   & Formalities at the beginning &  & 1 & 115 \\ \cline{2-5} 
(CO)        & Moderation &  & 1 & 90 \\ \cline{2-5} 
            & Impact factors & Analysis and clarification of problems & 13 & 873 \\ \cline{3-5} 
            &  & Analysis and agreement on objectives & 2 & 56 \\ \cline{3-5} 
            &  & Creating motivation & 4 & 131 \\ \cline{3-5} 
            &  & Resource activation & 5 & 117 \\ \cline{3-5} 
            &  & Help, problem solving & 8 & 71 \\ \cline{2-5} 
            & Formalities for conclusion and farewell &  & 2 & 60 \\ \cline{2-5} 
            & Other statements &  & 2 & 19 \\ \cline{2-5} 
            & \textbf{Total} &  & \textbf{38} & \textbf{1532} \\ \hline
Client      & Formalities at the beginning &  & 4 & 72 \\ \cline{2-5} 
(CL)        & Empathy &   & 3 & 95 \\ \cline{2-5} 
            & Impact factors & Analysis and clarification of problems & 8 & 437 \\ \cline{3-5} 
            &  & Analysis and agreement of objectives & 2 & 93 \\ \cline{3-5} 
            &  & Creating motivation & 2 & 75 \\ \cline{3-5} 
            &  & Resource activation & 2 & 63 \\ \cline{3-5} 
            &  & Help coping with problems & 6 & 282 \\ \cline{2-5} 
            & Formalities for conclusion &  & 2 & 70 \\ \cline{2-5} 
            & Other statements &  & 2 & 59 \\ \cline{2-5} 
            & \textbf{Total} &  & \textbf{28} & \textbf{1246} \\ \hline
\end{tabular}
}
\caption{Overview of the category hierarchy and label distribution.}\label{tab:CategoriesSamples}
\end{table*}

\subsection{Objectives}

Psychosocial online counseling is a professional support service, but assessing how established counseling factors \cite{grawePsychologischeTherapieGottingen2000} are reflected in large-scale text interactions requires scalable, fine-grained analysis beyond traditional resource-intensive methods.

As the basis for the analysis, a category system was developed according to the principles of qualitative social research \cite{mayringQualitativeContentAnalysis2015}. 
In the first step, analytical criteria were deductively derived from literature, primarily the key factors of effective counseling according to \cite{grawePsychologischeTherapieGottingen2000}, as well as other relevant structural elements of professional online counseling. 
For example, several categories derived from the theory of MI were integrated \cite{millerMotivationalInterviewingHelping2012}. 
Care was taken to ensure compatibility with the respective datasets mentioned in Section~\ref{sec:RelatedWork}, in order to maintain optimal comparability.
These individual elements were mapped to clear definitions, and typical anchor examples for corresponding text segments were established.

Using the first draft of the category system, suitable texts were annotated by specially trained human coders. In iterative team meetings, the results were compared and the category system was inductively developed further. 
For example, very similar categories were merged, or new categories were introduced to capture content that had not been adequately represented up to that point.


\subsection{Structure}
Based on these principles we created a new category system, consisting of 38 categories for counselor (CO) messages and 28 categories for client (CL) messages. The category system is hierarchically structured into at most five levels of detail. 
Table~\ref{tab:CategoriesSamples} shows the first two levels of the category system together with the respective number of categories at the lowest level (Level 5) and the number of annotated samples in the dataset. The names of all level-5 categories can be found in Table~\ref{tab:DetailedResults}.

With 38 different message types, the counselor side is more detailed than the client side. This is by design given that assessments of counseling quality and method adherence are driven foremost by by counselor messages. We use fine-grained categories for the most interesting type of messages for analysis, the so-called \textit{Impact Factors}. 

\begin{table*}[t]  
\centering
\small
\scalebox{0.95}{ 
\begin{tabular}{l l p{5cm} p{6cm} } \hline
\textbf{Category Code} &\textbf{Writer} & \textbf{Text} & \textbf{Category (Hierarchical Path)} \\ \hline

CO-IF-AC-RF-RLS-L & Counselor & To advise you better, it would be helpful to know what kind of hobbies or interests you have. &
Impact factors $\rightarrow$ Analysis and clarification of problems $\rightarrow$ Reflection (Fact) $\rightarrow$ 
Request about living situation $\rightarrow$ Leisure \\ \hline
CO-IF-AC-RF-RTP-* & Counselor & 
Can you describe the feeling you have afterwards a bit more closely? Is it more disgust or does it seem somehow wrong to you what you did? &
Impact factors $\rightarrow$ Analysis and clarification of problems $\rightarrow$ Reflection (Fact) $\rightarrow$ Targeted, precise request \\ \hline
CO-IF-HP-*-ITFE-* & Counselor & 
It is completely normal for a young person 
to start developing sexual desires and fantasies during puberty. & 
Impact Factors $\rightarrow$ Help, Problem Solving $\rightarrow$ Technical or factual explanations \\ \hline

CL-E-*-*-ECP-* & Client & It's more about my brother than about me. & Empathy $\rightarrow$ Concern for another person \\ \hline
CL-IF-ACP-*-FPA-* & Client & 
I have tried to alleviate my withdrawal symptoms with medication, but it feels like a constant struggle. &
Impact factors $\rightarrow$ Analysis and clarification of problems $\rightarrow$ Feedback on previous attempts  \\ \hline
\end{tabular}
}
\caption{Example utterances from counselors and clients and the respective category}\label{tab:ExampleUtterances}
\end{table*}

For example the impact factors at the counselor's level-2 subcategory \textit{Resource activation} are further distinguished into 
\textit{Question about possible support resources},
\textit{Request about problem statement}, 
\textit{Suggestion for activating resources at professional level},
\textit{Suggestion for activating resources at family level}, and
\textit{Suggestion for activating resources at friendship level}.
This kind of fine distinction between semantic categories allows for very detailed analysis.
However, many categories were defined at level four without further subdivision levels, because the level of detail was considered sufficient for the types of analyses we aim for. For example, all MI-related categories can be found on level 4. 

The highest resolution of five levels has the subcategory \textit{Impact factors $\rightarrow$ Analysis and clarification of problems}, because we want to analyze in detail the different types of questions and methods which counselors use to get to the core of the client's problem. Some examples can be found in Table~\ref{tab:ExampleUtterances}.

We created unique short codes for each of the categories for simpler data processing. Each category at each level is assigned such a short code (values in the parentheses in the example above). The complete code of a category at lower level in the hierarchy is the concatenation of codes along the whole path, i.e. for \textit{Social relationships} it is \textit{CO-IF-AC-RF-RLS-SR}. These codes uniquely identify the categories at the finest level of detail and form the labels we trained the classifiers on.


Our categories can be seen as a domain‑specific extension of broader dialogue‑act tagging frameworks (e.g., ISO 24617‑2). 
At a coarse level, categories such as \textit{Formalities} correspond to \textit{Social Obligations Management} (e.g., greeting, closing), \textit{Moderation} corresponds to \textit{Interaction Management} (e.g., turn/grounding management), and parts of \textit{Impact factors} map to \textit{Task} and \textit{Feedback} functions (e.g., information-seeking, advising, checking understanding), while refining them into counseling-specific strategies (e.g., resource activation, motivation building). This positioning facilitates comparison to prior work using general dialogue-act standards while preserving the counseling-specific granularity of OnCoCo.

\section{The OnCoCo 1.0 Dataset}

To enable a fine-grained analysis of interaction patterns between counselors and clients, we created a new bilingual dataset (German with English translations).

\subsection{Curation and Annotation}

The creation of the dataset is a critical step in ensuring the validity and applicability of the fine-grained message classification system for online counseling conversations. Particular care was taken to ensure privacy, ethical integrity, and data quality. 

\paragraph{Origin of Data:}
The research team deliberately refrained from anonymization existing real-world counseling datasets, as these either lacked explicit consent for research use or involved ethical concerns regarding the use of publicly available counseling content \cite{ghanemBigDataUnd2022, bruckmanStudyingAmateurArtist2002}. 
Instead, the social scientists took inspiration from real data in public discussion forums and educational role-play sessions. They also drew on their extensive prior research experience to develop synthetic interactions that closely mirror the linguistic patterns, emotional dynamics, and structural characteristics of real counseling exchanges.
In multiple iterative steps, the generated sample texts were evaluated for authenticity and compared to real data in order to ensure a high degree of realism and representativeness.
In addition, the data was carefully curated and reviewed to minimize potential biases and avoid the reinforcement of stereotypes. 


\paragraph{Privacy and Ethics:} 
As the data were generated by experts, no personal data are contained in the dataset. Thus, no further ethical approval was required to publish the data.


\paragraph{Annotation Process:}
\cite{scheuermanHowWeveTaught2020} rightly point out that the annotation of datasets can introduce unconscious biases into classification processes. This effect can never be entirely ruled out. The annotation and category development in this study were carried out by individuals with the following identities and backgrounds: 

The annotation team consisted of six trained online counselors and five student assistants, including students of social work and computer science. All annotators were born in Germany and held a university entrance qualification. The group included 9 women and 2 man.

Thus, on the one hand, it cannot be ruled out that the lack of diversity among the annotators may have introduced implicit bias into the dataset. On the other hand, the bachelor’s program in social work in Germany includes targeted training designed to help students recognize and reflect on their own biases and engage in their professional work with cultural humility and awareness.

The annotators were instructed in detail in the underlying methodology of the category system. To ensure high data quality in this complex social domain, the labels were established through expert consensus. Annotators discussed disagreements in regular meetings until a shared interpretation of the 66 categories was reached. Final labels and category definitions were reviewed by a lead expert in online counseling to ensure theoretical validity and consistent use across the hierarchy. 
All student assistants had previous experience with the annotation of counseling data within the field of psychosocial counseling. This whole group was professionally guided by the project coordinator and the project management.

\begin{table*}[htb]  
\centering
\small
\scalebox{0.95}{ 
\begin{tabular}{|c|c|c|c|c|c|}
\hline
Model    & Parameters         & \multicolumn{2}{c|}{Accuracy} & \multicolumn{2}{c|}{F1 Macro} \\ 
\cline{3-6} 
                              &   & @1   & @2   & @1   & @2   \\ \hline
FacebookAI/xlm-roberta-large   & 561M & \textbf{0.79} $\pm$0.02 & \textbf{0.88} $\pm$0.01 & \textbf{0.72} $\pm$0.02 & \textbf{0.83} $\pm$0.02 \\ 
FacebookAI/xlm-roberta-base    & 279M & 0.76 $\pm$0.02 & 0.86 $\pm$0.01 & 0.68 $\pm$0.03 & 0.79 $\pm$0.02 \\ \cline{1-6}
answerdotai/ModernBERT-large   & 396M & 0.77 $\pm$0.01 & 0.86 $\pm$0.01 & 0.69 $\pm$0.01 & 0.80 $\pm$0.02 \\
answerdotai/ModernBERT-base    & 150M & 0.68 $\pm$0.02 & 0.78 $\pm$0.00 & 0.57 $\pm$0.02 & 0.69 $\pm$0.01 \\ \cline{1-6}
EuroBERT/EuroBERT-610m         & 610M & 0.76 $\pm$0.01 & 0.84 $\pm$0.01 & 0.69 $\pm$0.02 & 0.79 $\pm$0.02 \\
EuroBERT/EuroBERT-210m         & 210M & 0.75 $\pm$0.01 & 0.84 $\pm$0.01 & 0.67 $\pm$0.02 & 0.79 $\pm$0.03 \\ \hline 
\end{tabular}
}
\caption{Classification results for selected fine-tuned models on cross validation data,
showing mean and sample standard deviation across five folds and second best choice (@2).}\label{tab:ClassificationResults}
\end{table*}

\paragraph{Translation:} 
All conversations were originally written in German and then automatically translated into English. The translation was done using GPT-4o. A review based on random samples confirmed the expected high quality of the translations. 

\subsection{Description of the Dataset}
The actual dataset comprises a total of 2,778 original messages. The data distribution is skewed with regard to the categories (Table~\ref{tab:CategoriesSamples}). This is due to the naturally heterogeneous distribution of messages in counseling dialogues. 
Most of the client and counselor data belongs to the respective top-level category \textit{Impact factors}, as this is the focus of methodological analysis. Table~\ref{tab:ExampleUtterances} gives an impression of what the text examples look like.





\begin{table*}[h]  
\renewcommand{\arraystretch}{0.9}
\small
\centering
\scalebox{0.8}{ 
\begin{tabular}{|l|l|r|r|r|r|}
\hline
Code & Level 5 & Precision & Recall & F1 Score & Support \\ \hline 
CO-FA-*-*-*-* & Formalities at the beginning/Opening of the conversation & 0.90 & 0.96 & 0.93 & 46 \\
CO-FC-*-*-F-* & Farewell & 1.00 & 0.95 & 0.97 & 20 \\
CO-FC-*-*-OPR-* & Offer to use the counselor's professional resources & 0.67 & 1.00 & 0.80 & 4 \\
CO-IF-AC-RE-RCR-* & Complex Reflection & 0.50 & 0.67 & 0.57 & 6 \\
CO-IF-AC-RE-RES-* & Request for emotional state & 0.68 & 0.72 & 0.70 & 18 \\
CO-IF-AC-RF-RC-* & Request for concerns & 0.63 & 0.79 & 0.70 & 24 \\
CO-IF-AC-RF-RCD-* & Request for change and development & 0.86 & 0.40 & 0.55 & 30 \\
CO-IF-AC-RF-RLS-ES & Employment/economic situation & 0.42 & 0.42 & 0.42 & 12 \\
CO-IF-AC-RF-RLS-H & Health (mental and physical) & 0.83 & 0.67 & 0.74 & 30 \\
CO-IF-AC-RF-RLS-L & Leisure & 0.82 & 0.82 & 0.82 & 34 \\
CO-IF-AC-RF-RLS-PS & Professional situation & 0.70 & 0.82 & 0.76 & 34 \\
CO-IF-AC-RF-RLS-SR & Social relationships & 0.72 & 0.91 & 0.81 & 34 \\
CO-IF-AC-RF-RPA-* & Request for previous attempts at solutions & 0.88 & 0.91 & 0.89 & 32 \\
CO-IF-AC-RF-RPD-* & Request for personal data & 0.89 & 0.61 & 0.72 & 28 \\
CO-IF-AC-RF-RTP-* & Targeted, precise request & 0.39 & 0.45 & 0.42 & 20 \\
CO-IF-AC-RF-SRx-* & Simple reflection & 0.94 & 0.96 & 0.95 & 52 \\
CO-IF-AO-*-ICO-* & Definition of counseling objectives & 0.64 & 0.58 & 0.61 & 12 \\
CO-IF-AO-*-ROW-* & Request for objectives/wishes & 0.37 & 0.70 & 0.48 & 10 \\
CO-IF-HP-*-ICO-* & Calming & 1.00 & 1.00 & 1.00 & 2 \\
CO-IF-HP-*-IEA & Evaluation, Interpretation & 0.60 & 1.00 & 0.75 & 6 \\
CO-IF-HP-*-IF-* & Future forecast & 0.00 & 0.00 & 0.00 & 2 \\
CO-IF-HP-*-IPFR-* & Professional, formal, institutional recommendation & 1.00 & 1.00 & 1.00 & 2 \\
CO-IF-HP-*-ITFE-* & Technical or factual explanations & 1.00 & 1.00 & 1.00 & 6 \\
CO-IF-HP-*-IW-* & Warning & 1.00 & 1.00 & 1.00 & 4 \\
CO-IF-HP-*-PP-IA & Advice & 0.67 & 1.00 & 0.80 & 2 \\
CO-IF-HP-*-PP-IW & Wish & 1.00 & 1.00 & 1.00 & 4 \\
CO-IF-Mot-*-IAC-* & Articulation of the perceived ability to change in the client (MI) & 1.00 & 0.70 & 0.82 & 10 \\
CO-IF-Mot-*-IEM-* & Encouragement, motivation of the client & 0.82 & 1.00 & 0.90 & 14 \\
CO-IF-Mot-*-ITA-* & Thanks and appreciation to the client & 1.00 & 0.86 & 0.92 & 14 \\
CO-IF-Mot-*-RFC-* & Eliciting/Evoking "Change-Talk" (MI) & 0.80 & 0.57 & 0.67 & 14 \\
CO-IF-Mot-*-RS-* & Question about possible support resources & 0.64 & 0.64 & 0.64 & 14 \\
CO-IF-RA-*-N-RAFa & Suggestion for resource activation at the family level & 1.00 & 1.00 & 1.00 & 10 \\
CO-IF-RA-*-N-RAFr & Suggestion for resource activation at the friendship level & 1.00 & 1.00 & 1.00 & 2 \\
CO-IF-RA-*-RAP-* & Suggestion for resource activation at professional level & 1.00 & 1.00 & 1.00 & 14 \\
CO-IF-RA-*-RP-* & Request for problem statement & 1.00 & 0.67 & 0.80 & 6 \\
CO-Mod-*-*-*-* & Moderation & 0.81 & 0.81 & 0.81 & 36 \\
CO-O-*-*-O-* & Other statements & 0.00 & 0.00 & 0.00 & 2 \\
CO-O-*-*-UCO-* & Inappropriate remark & 1.00 & 0.50 & 0.67 & 6 \\ \hline 
CL-E-*-*-ECC-* & Compassion for others (EC) & 0.83 & 0.71 & 0.77 & 14 \\
CL-E-*-*-ECP-* & Concern for another person (EC) & 0.61 & 0.92 & 0.73 & 12 \\
CL-E-*-*-PT-* & Empathy for third parties or related to the present situation (PT) & 0.78 & 0.58 & 0.67 & 12 \\
CL-FB-*-*-*-* & Formalities at the beginning and greeting & 1.00 & 1.00 & 1.00 & 28 \\
CL-FC-*-*-F-* & Formalities for conclusion & 1.00 & 1.00 & 1.00 & 8 \\
CL-FC-*-*-UPR-* & Further use of professional resources of counselor & 0.83 & 0.95 & 0.88 & 20 \\
CL-IF-ACP-*-Cons-* & Consent & 0.78 & 1.00 & 0.88 & 14 \\
CL-IF-ACP-*-DPD-* & Disclosure of personal data & 1.00 & 0.95 & 0.97 & 20 \\
CL-IF-ACP-*-FPA-* & Feedback on previous attempts at solutions & 0.88 & 0.85 & 0.86 & 26 \\
CL-IF-ACP-*-OE-* & Own emotional expression & 0.95 & 0.95 & 0.95 & 22 \\
CL-IF-ACP-*-PD-* & Problem definition & 0.44 & 0.79 & 0.56 & 14 \\
CL-IF-ACP-*-PS-* & Problem statement & 0.72 & 0.70 & 0.71 & 30 \\
CL-IF-ACP-*-Rej-* & Rejection & 0.85 & 0.85 & 0.85 & 20 \\
CL-IF-ACP-*-Req-* & General request & 1.00 & 0.93 & 0.96 & 28 \\
CL-IF-AO-*-Ext-* & Extension of the assignment & 0.83 & 0.36 & 0.50 & 14 \\
CL-IF-AO-*-Obj-* & Objective of the assignment & 0.67 & 0.83 & 0.74 & 24 \\
CL-IF-HP-*-Fail-* & Final failure & 0.93 & 0.72 & 0.81 & 18 \\
CL-IF-HP-*-NegFR-* & Negative feedback on specific action recommendation & 0.68 & 0.77 & 0.72 & 22 \\
CL-IF-HP-*-PosF-* & General positive feedback & 0.83 & 0.94 & 0.88 & 16 \\
CL-IF-HP-*-PosFR-* & Positive feedback on specific recommendations for action & 1.00 & 0.88 & 0.94 & 26 \\
CL-IF-HP-*-RepRA-* & Report on the implementation of recommendations for action & 0.91 & 0.83 & 0.87 & 12 \\
CL-IF-HP-*-Succ-* & Final success & 1.00 & 0.89 & 0.94 & 18 \\
CL-IF-Mot-*-FC-* & Eliciting/Evoking "Change-Talk" (MI) & 0.53 & 0.67 & 0.59 & 12 \\
CL-IF-Mot-*-RC-* & Articulation of reasons for a change in the client & 0.80 & 0.67 & 0.73 & 18 \\
CL-IF-RA-*-RF-* & Considering resource activation at the level of friends and family & 0.89 & 1.00 & 0.94 & 8 \\
CL-IF-RA-*-RP-* & Considering resource activation at a professional level & 0.73 & 0.50 & 0.59 & 16 \\
CL-O-*-*-O-* & Other statements & 1.00 & 0.50 & 0.67 & 8 \\
CL-O-*-*-UCO-* & Inappropriate remark & 0.88 & 0.94 & 0.91 & 16 \\ \hline
\end{tabular}
}
\caption{Classification report for the best performing model xlm-roberta-large on the 1,112 test samples}\label{tab:DetailedResults}
\end{table*}

\section{Classification}

The dataset in its current form is created for classification. 
The related work described in Section~\ref{sec:RelatedWork} showed that transformer-based models are generally well suited for this purpose. However, existing data sets use just a few classes while our category system consists of 38 classes for counselor messages and 28 for clients. Multi-class classification with a high number of classes is always a challenge, especially if some classes are semantically close like \textit{"Generation of motivation"} and \textit{"Resource activation"}. The challenge becomes even more difficult if the number of records per category is low.

\subsection{Experimental Setup}

To demonstrate the usefulness of our data for classification, we chose a set of available multi-lingual encoder models and performed full fine-tuning for classification on our dataset (see Table~\ref{tab:ClassificationResults}). Our first choice fell on XLM RoBERTa which has shown high performance on numerous classification tasks. In addition, we included the two more recently publish model families ModernBERT \cite{warnerSmarterBetterFaster2024} and EuroBERT \cite{boizardEuroBERT2025}. For each of these, we fine-tuned a smaller base model and a larger variant.

We experimented with separate models for counselors and clients, since the author of a message is known apriori and does not require classification. It turned out, however, that a single model trained on both types of messages is on par with separate models if the messages are simply prefixed with either "Counselor:" or "Client:".
One drawback of the unified model is the potential for confusion between counselor and client categories during inference. To address this, we applied output masking by suppressing client-specific categories for counselor messages and vice versa at the level of the prediction logits. We also experimented with applying this masking during training, but observed no measurable improvement in performance.


\subsection{Experimental Results}

In order to report robust results, we performed 5-fold cross-validation with each model. 
Table~\ref{tab:ClassificationResults} gives an overview of the results.
We report the mean and sample standard deviation of the evaluation metrics across the validation data of the five folds.
Our primary evaluation metric was the F1 Macro score, i.e. the unweighted average of the F1 score per class because the mean accuracy gives little indication of possibly low recall of classes with little support.
As some categories are semantically very close and even for humans hard to distinguish, we also evaluated the second-best choice of the model (Accuracy@2 and F1@2). 

The best performing model, XLM RoBERTa, reaches an accuracy of 0.79 and an F1 Macro of 0.72, increasing by over 10 points to 0.88 and 0.83 by taking also the second best prediction into account.
Most categories could be detected with high recall and even categories with very little training data are recognized relatively well by this model, as can be seen in Table~\ref{tab:DetailedResults}. 
ModernBERT and EuroBERT also perform well, but lag behind by a few points in both accuracy and F1 score. There is also a measurable performance degradation if smaller base models are used.

Beyond quantitative metrics, a qualitative examination of the outputs of the best performing model reveals that our OnCoCo dataset successfully captures the core interaction structures of professional online counseling. The models perform strongly in identifying categories related to problem clarification, empathy, and resource activation—key mechanisms known from counseling research to contribute to effective psychosocial support \cite{grawePsychologischeTherapieGottingen2000,eichenbergZurWirksamkeitOnlineBeratung2016}. The classification behavior generally aligns with theoretical expectations: counselor messages emphasizing clarification, motivational support, and structured counseling display high internal coherence and distinct linguistic markers, while categories that rely on subtle affective or contextual cues remain more challenging. This suggests that the OnCoCo category system and trained models are capable of reproducing the interpretive logic of human coders and, therefore, offer a bridge between qualitative content analysis and computational modeling in social work research.

Categories with low F1 scores in Table~\ref{tab:DetailedResults} mainly reflect semantic overlap and data sparsity. For example, the client categories \textit{Problem statement} and \textit{Problem definition} are often confused because both describe aspects of clients' difficulties, differing only in abstraction level. Similarly, the counselor categories \textit{Targeted, precise request} and \textit{Request for change and development} show pragmatic similarity, leading to misclassification. 
In these cases, the second best (@2) choice of the model often matches the label. 
Rare types such as \textit{Complex reflection} or \textit{Future forecast} suffer from too few examples for stable learning. Motivational categories like \textit{Eliciting/Evoking Change-Talk} also remain challenging, as such utterances are short and context-dependent. Overall, explicit or formulaic acts (e.g., greetings, farewells) are classified reliably, while categories requiring pragmatic inference or emotional interpretation yield lower performance—mirroring known limits of text-only models in capturing counseling subtleties.

\subsection{Intercoder Reliability}


\paragraph{Human-to-human:} 
To assess the reliability of the annotation process, we conducted a systematic inter-annotator agreement (IAA) study. All items from the dataset were independently coded by two human annotators from the trained annotation team, covering all categories of both message roles. The Macro-averaged Cohen's $\kappa$ across all items is 0.84 (0.85 for counselor and 0.80 for client utterances) and can be considered very strong. 
As expected in fine-grained counseling dialogue, agreement was higher for structural categories (e.g., \textit{Greetings}, \textit{Formalities}) and more moderate for semantically dense \textit{Impact Factors} such as \textit{Resource Activation}. However, the overall Macro-averaged $\kappa$ of 0.84 indicates that the coding scheme remains robust and learnable across these varying levels of complexity.
Gold-standard labels were finally established through expert consensus: disagreements were discussed between the annotators and a lead expert in psychosocial online counseling until a shared interpretation was reached.

\paragraph{Human-to-model:}
We additionally assessed the agreement between the best-performing model (XLM-RoBERTa-large) and codings from human experts. A stratified sample of 20\% of the original German utterances (N=556), evaluated against the same gold-standard labels as the classification task, yielded a Cohen's $\kappa$ of 0.88. This high value strongly indicates that the trained model's predictions are conceptually coherent with human expert interpretation. As summarized in Table~\ref{tab:DetailedResults}, while the cross-validated Macro F1 (0.72) is numerically lower, largely due to the metric's high sensitivity to rare categories in a 66-class set, the proximity between the model's performance and the initial human inter-annotator agreement ($\kappa=0.84$) demonstrates that the automated classification is approaching the performance ceiling for this granular task.

\section{Summary and Further Research}



This paper introduced OnCoCo 1.0, a bilingual, fine-grained dataset and category system for the automated analysis of psychosocial online counseling conversations. Integrating diverse counseling approaches, the system enables the detailed classification of counselor-client interactions and supports scalable, data-driven evaluation of counseling practices. 
The created OnCoCo dataset, together with fine-tuned classification models, offers significant potential for improving the quality and efficiency of psychosocial support services. 

While the models already achieve satisfactory results for most categories, certain fine-grained types remain difficult to distinguish due to semantic overlap, subtle contextual cues, and limited data support. Future research should focus on enhancing the modeling of pragmatic and affective dimensions, for instance by integrating discourse-level features or multi-task learning objectives, to strengthen the classification of conceptually overlapping categories. Further improvements could also result from expanding the dataset with additional annotated examples for underrepresented categories or by leveraging large language models to generate high-quality synthetic samples. 


\newpage
\section{Limitations}

Our dataset and the models come with several limitations:
\paragraph{Skewed Data Distribution:} The dataset has an uneven distribution of categories, which may lead to challenges in detecting underrepresented categories. While it is planned expand the dataset, the current imbalance affects model performance.
\paragraph{Potential Generalization Issues:} Although the models achieve high accuracy and F1, the performance may drop on data coming from types of online counseling which were not covered by the training data.
\paragraph{Limited Cultural Adaptability:} The dataset, though bilingual (German and English), does not explicitly address cultural differences in counseling or how these might affect the applicability of the models across diverse populations and languages.
\paragraph{Narrow Scope of Application:} While the dataset and models are focused on online counseling, their application is limited to this domain. Expanding the models to broader contexts, such as peer support or general social work, would enhance their societal impact.
\paragraph{Sociolinguistic and Ethical Limitations:}
From a qualitative and sociological perspective, the dataset represents online counseling dialogues under idealized professional conditions. This focus ensures ethical soundness and methodological consistency, but also constrains the variety of linguistic and relational phenomena captured. Informal, ambiguous, or boundary-crossing communication (e.g. such as humor expressions, self-disclosure, or other indirect emotional cues) is rare or even absent. While this reflects the professional standards of counseling practice, it can reduce the ability of the model to generalize to less formal contexts, where such elements are relevant to building rapport. In addition, the synthetic, but expert-generated nature of the dialogues means that spontaneous and situational nuances of real counseling interactions are only partially comprised. 


\vspace{2cm}

\section{Bibliographical References}\label{sec:reference}

\bibliographystyle{lrec2026-natbib}
\setcitestyle{maxcitenames=3, maxbibnames=3}
\bibliography{lrec2026-example}


\end{document}